\documentclass[letterpaper]{article} 
\usepackage{times}  
\usepackage{helvet}  
\usepackage{courier}  
\usepackage[hyphens]{url}  
\usepackage{graphicx} 
\usepackage{natbib}  
\usepackage{caption} 
\usepackage{amsmath}
\usepackage{amsthm}
\usepackage{multirow}
\usepackage{booktabs}
\usepackage{graphicx}
\usepackage{amssymb}
\usepackage{mathrsfs}
\usepackage{subcaption}
\usepackage{dsfont}
\usepackage{xcolor}

\usepackage{algpseudocode}

\theoremstyle{definition}

\newtheorem{theorem}{Theorem}

\begin{document}
%
\title{FedASTA: Federated Adaptive Spatial-Temporal Attention for Time Series Prediction}
\author{Kaiyuan Li \thanks{likaiyua23@mails.tsinghua.edu.cn}, Yihan Zhang \thanks{zyh23@mails.tsinghua.edu.cn}, Huandong Wang, Yan Zhuo, Xinlei Chen \\
Tsinghua University
}
\maketitle
\begin{abstract}
Mobile devices and Internet of Things (IoT) devices generate large amounts of spatiotemporal data. Predicting spatiotemporal data while preserving privacy remains a challenging problem. Federated learning-based distributed time series prediction frameworks have been proposed as a solution, reducing privacy concerns by not sharing original data. Traditional methods often rely on model parameter exchange, but this approach overlooks the spatial relationships between nodes. Recently, approaches based on graph neural networks have been proposed to capture spatial relationships between different nodes, to model complex spatiotemporal relationships. However, existing methods only consider static distance relationships between nodes and rarely account for dynamic spatiotemporal relationships.
To address these challenges, we propose a novel Federated Adaptive Spatial-Temporal Attention (FedASTA) framework. We define a new distance calculation method in the frequency domain to compute the similarity between temporal data. Based on this calculation method, we efficiently construct an adaptive spatiotemporal graph. We then propose a new mask attention mechanism based on the static distance graph and the dynamic spatiotemporal graph, effectively modeling the dynamic spatiotemporal relationships between different nodes. Additionally, we discuss the privacy and security issues faced by this framework and propose a protection scheme.
Extensive experiments on six real-world public datasets demonstrate that our method achieves state-of-the-art performance in federated scenarios. Studies on model scalability, communication efficiency, robustness, and ablation studies on each module further prove the effectiveness of the model.
\end{abstract}

\section{Introduction}

Nowadays, mobile devices and Internet of Things (IoT) devices generate a substantial volume of spatiotemporal data, which has garnered significant attention in academic research on spatiotemporal data modeling. Most existing studies assume that a centralized model can access the entire dataset \cite{wu2021autoformer,lan2022dstagnn,jiang2021dl}. However, this assumption may not hold for spatiotemporal data: electricity consumption data can potentially reveal sensitive information about whether a user is at home \cite{apthorpe2017spyingsmarthomeprivacy} , while traffic flow data is typically collected through cameras and controlled by various organizations \cite{9082655,liu2023onlinespatiotemporalcorrelationbasedfederated} , such as government agencies, taxi companies, and ride-hailing platforms. Due to privacy considerations, data is not typically shared between different organizations. Federated learning (FL) \cite{mcmahan2017communication}, as a machine learning framework, was introduced to handle these issues.

\begin{figure}[t]
    \centering
    \includegraphics[width=0.9\linewidth]{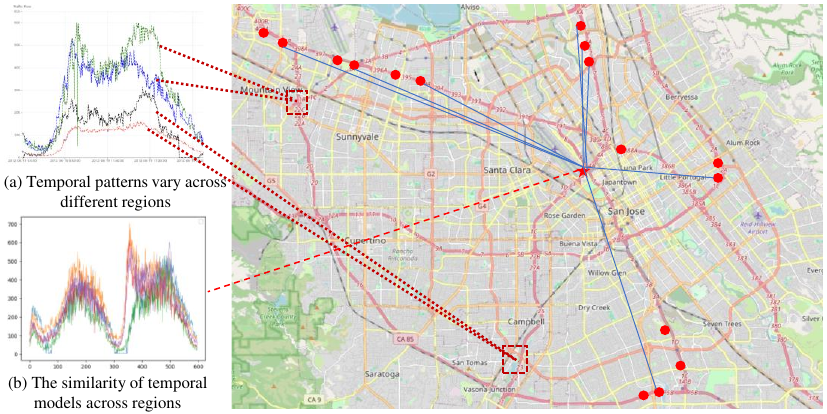}
\caption{(a) Different regions exhibit distinct temporal features.
(b) Sensors with similar temporal features are not necessarily spatially adjacent.}
\vspace{-16pt}
\end{figure}


Traditional federated learning methods \cite{li2020federated}, \cite{9488883},\cite{pmlr-v139-collins21a} have achieved training on distributed data across distributed devices. However, these methods often rely on sharing model parameters or gradients and regularizing the model parameters across different nodes to learn a model that fits the data distribution of all nodes. These approaches do not consider the inherent spatial and temporal relationships between nodes in spatiotemporal data. As shown in Figure 1(a), we selected nodes from different locations within two regions. It is evident that nodes located close to each other within the same region exhibit similar characteristics, whereas the characteristics between nodes from different regions vary significantly.

Recently, some federated learning methods have begun to focus on the spatiotemporal relationships between nodes. FedDA \cite{9488883} groups clients into different clusters by using a small augmented dataset to extract temporal features. However, this approach does not take into account the intrinsic positional relationships between nodes.   CNFGNN \cite{meng2021cross} introduces a static distance graph and models the spatial relationships between different nodes using a Graph Neural Network (GNN). But the spatiotemporal relationships between nodes often change dynamically over time. As shown in Figure 1(b), static physical distance cannot effectively model the spatiotemporal relationships between different locations. For example, traffic flow in the same urban functional areas, such as school zones during school hours, exhibits similar temporal patterns even if they are not geographically close. Currently, methods that dynamically compute node relationship graphs based on attention mechanisms \cite{jiang2023pdformer} often assign excessive weight to irrelevant nodes, while also incurring significant computational overhead.
To address above problems, in this paper, we propose \textbf{Fed}erated \textbf{A}daptive \textbf{S}patial-\textbf{T}emporal \textbf{A}ttention Learning (FedASTA), an FL framework designed to model the complex spatial-temporal dynamics. On the client side, we use an encoder and Fourier transform to process the seasonal and trend components obtained through the time series decomposition module. The client transfer the temporal features to server node. On the server side, we propose a novel adaptive spatiotemporal relation graph construction module to capture the dynamic spatiotemporal relations. Based on this module, we further design a mask attention mechanism to simplify computation and constrain the model's attention on the right nodes. Using a static distance graph mask constructed based on the distances between nodes and a dynamic spatiotemporal relationship graph mask built by the adaptive graph construction module, server-side nodes can more accurately integrate client node features according to spatiotemporal relationships. Finally, we analyze potential privacy issues in the model architecture and provide protection solutions.
In summary, the main contributions of our work are as follows:

\begin{enumerate}
    \item We propose a novel distance calculation method to support dynamic computation of connections between nodes and further design an adaptive spatiotemporal graph construction module to capture complex spatial-temporal dynamics among nodes.
    \item The proposed sparse attention mechanism greatly reduce computational overhead while achieves comparable performance with the complete one.
    In addition, we analyzed the privacy issues and proposed protection solutions.
    \item Extensive experiments on six real-world datasets show that our method can outperform other FL baselines. The visualization experiments enhanced the interpretability of our proposed graph construction module.
\end{enumerate}

\section{Related work}
\paragraph{Federated Learning} Traditional federated learning methods \cite{mcmahan2017communication,li2017diffusion,pmlr-v139-collins21a} typically perform fusion operations on model parameters or gradients to train a global model. With the increasing volume of edge data, using federated learning to predict spatiotemporal data at the edge has garnered widespread research interest. Graph neural network-based FL methods are commonly used to model the correlations between nodes \cite{meng2021cross,yuan2022fedstn,qin2024spatio}. FedGRU \cite{9082655} propose an ensemble clustering-based method, client nodes are grouped into different categories for aggregation. FedDA \cite{9488883} utilize dual attention-based module to predict wireless traffic flow. 
Although this work addresses static distance spatial relations, it neglects the dynamic spatial-temporal relations among client nodes. 

\paragraph{Dynamic construction of graph}
Recent works have focused on constructing dynamic relation graphs to model spatial-temporal relations among nodes. GMAN \cite{zheng2020gman} maps node distances to edge weights. Learning based methods like \cite{guo2019attention,zhang2020spatio,bai2020adaptive} use additional learnable parameters for dynamic relations but lack interpretability and require large data, limiting their use in federated learning. STSGCN \cite{song2020spatial} uses a simple binary graph with limited characterization, considering only the nearest time steps. DSTAGNN \cite{lan2022dstagnn} dynamically calculates node distances using the Wasserstein distance, but this method is time-consuming and requires the full time series. 
Our work aims to propose a more computationally efficient and interpretable method for dynamic graph construction.


\section{Methodology}


In this section, we first introduce the overall framework. Then, we describe in detail the client model and the masked attention module with adaptive graph construction module on the server side. Finally, we analyze the privacy considerations of the overall framework and provide a privacy protection mechanism. The formal representation of the complete process on both sides is given in Appendix E and F respectively.

\subsection{Problem Formulation}
A sensor network can be abstracted as a graph $\mathcal{G} = (\mathcal{V},\mathcal{E})$, where $\mathcal{V} = \{v_1, \cdots ,v_N\}$ represents the set of $N$ sensors, and $\mathcal{E}$ is the set of edges, whose weights represent the degree of correlation between the data at different sensors. We use \(X_i \in \mathbb{R}^{L_i \times D}, i \in [N] \) to denote the data of sensor \(i\), in which \(L_i\) is the number of time steps and $D$ is the feature dimension. 
Each node \(i\) has its own edge model \(\mathcal{M}_i\) and a central model \(\mathcal{M}_{central}\) is required for necessary information communication among nodes is designed to model the spatial-temporal dynamics. We represent the information node \(i\) shares with central model as \(h_i\) and the merged message got from central model as \(h_{agg_i}\). Formally, our goal is to use edge model \(\mathcal{M}_i\) with sharing message \(c_i\) to predict \(y_i\),
\begin{equation}\label{d}
    [h_{1,\mathrm{agg}},\cdots,h_{N,\mathrm{agg}}] = \mathcal{M}_{central}([h_1,\cdots,h_N];\mathcal{G}),
\end{equation}
\begin{equation}\label{d}
    y_i = \mathcal{M}_i(x_i;h_{i,\mathrm{agg}};h_i).
\end{equation}

\begin{figure*}[h]
    \centering
    \includegraphics[width=\linewidth]{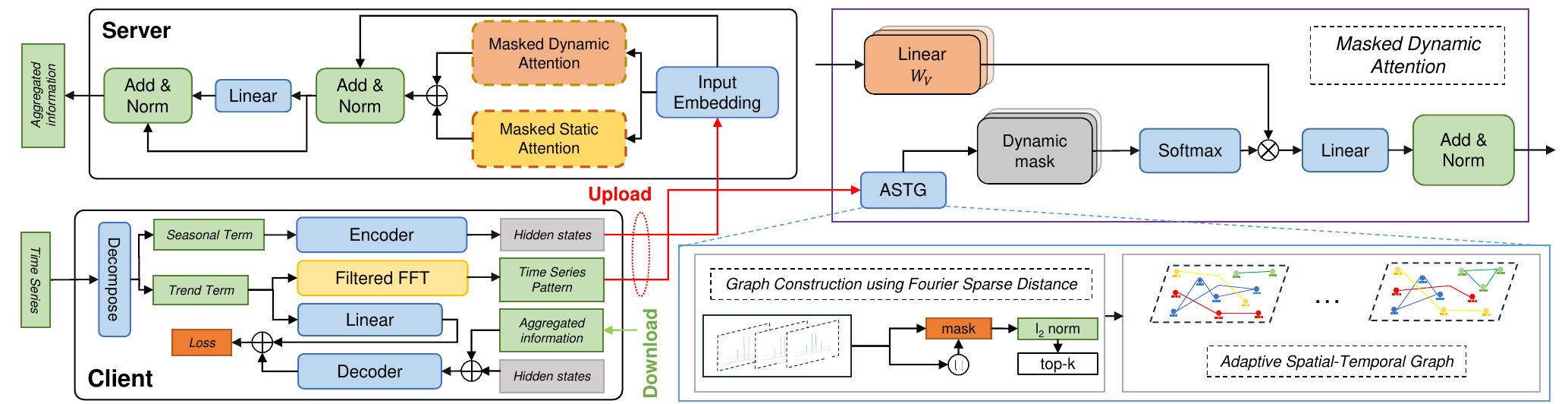} 
    \caption{Overall Structure}
    \label{fig:overall}
\vspace{-10pt}
\end{figure*}


As shown in Figure \ref{fig:overall}, our framework consists of a client side and a server side. The client side utilizes an encoder to compute its temporal features \(h_i\) and employs a Filtered FT module to extract trend features \(h_{trend}\) from the trend component. The client then uploads these two features to the server side. The server side’s adaptive graph construction module (ASTG) uses the trend features uploaded by the client to calculate the dynamic spatiotemporal relationship graph among client nodes. Subsequently, the static distance graph and the dynamic spatiotemporal relationship graph are utilized to aggregate the temporal features of spatiotemporally related nodes through a mask attention module. Finally, the server distributes the aggregated features back to the client, where the client combines its own features with the aggregated features for prediction.

\subsection{Client Model}

For each client model, it is necessary to model the self-temporal dynamics and extract temporal trend features for adaptive graph construction, as mentioned before. However, directly extracting trend from the original time series, which often has complex patterns, can be challenging and time-consuming. 

So we first utilize time series decomposition module to decompose time series into seasonal part and trend part on each client \(i\). 
The time series decomposition module uses average pooling (Avgpool) to smooth the original series and obtain the trend part. Then, the seasonal part is obtained by subtracting the trend part from the original series. 
Then, we use a GRU-based encoder to extract the temporal features \(h_i\) of each node \(i\) from the seasonal part. To represent the trend feature \(trend_i\) with only a small set of features, we apply a Filtered Fourier Transform which will be detailed in the next section to the trend term and obtain a representation of signals in the frequency domain. After obtaining both features, clients upload them to the server.

After receiving aggregated temporal-spatial information \(h_{i,agg}\) from the central server node, the client combines its own temporal dynamics \(h_i\) with the aggregated information \(h_{i,agg}\) to get seasonal prediction. Finally, we utilize a linear layer to project the trend part and sum up the seasonal and trend predictions to obtain the final prediction result \(\hat{y}_i\).

\subsection{Adaptive Spatial-Temporal Graph Construction}
In this section, we introduce our module for adaptive temporal-spatial graph construction, enabling dynamic construction on large-scale datasets in a computationally feasible way.
Our adaptive graph captures dynamic temporal-spatial relations rather than static distance relations.
Unlike the approach presented in \cite{lan2022dstagnn}, we do not calculate similarity using the entire sequence, as it is extremely time-consuming. Inspired by strong periodicity of time series, we leverage the Fourier Transform (FT) to develop a more efficient algorithm that achieves better performance.

With regard to the traffic flow sequence \(X_i\) in a node \(i\), we first apply Discrete Fourier Transform (DFT),
\begin{equation}
    \Lambda_i=\mathcal{F} \{X_i\},
\end{equation}
where \(\Lambda_i\) is the representation of \(X_i\) in the frequency domain, containing two components: amplitude and phase. It is of the same length as \(X_i\). 
Its elements are in the complex field, that is \(\Lambda_i\in \mathbb{C}^T\), with the corresponding frequencies
\(Q_i=\Big(0,1,\dots,\left \lfloor \frac{N-1}{2} \right \rfloor ,\) \(-\left \lfloor \frac{N}{2} \right \rfloor,\dots,-1\Big)\).
Then we apply sparsification to \(\Lambda_i\). Specifically, we set a threshold \(\mu\) and retain components with module length larger than \(\mu\) discarding the others. This process can be represented as follows:
\begin{equation}
    \Lambda_{i,\mu}(k) = \Lambda_i(k) \cdot \mathds{1}(|\Lambda_i(k)| > \mu),
\end{equation}
where \(k\) denotes the \(k\)th component.
The advantage of this step is 
We refer to the selected components as the Fourier main components and the process as the Filtered FT. This also results in an index set for each client \(i\),
\begin{equation}
    H_{i}=\{k|\Lambda_{i,\mu}(k)\neq 0\}.
\end{equation}
Note that the size of \(H_i\) may vary for different client \(i\). We denote the union of the bases of different clients as \(\mathbf{H}_u=\bigcup_{i=1}^{N}H_{i}\), and \(T_u=|\mathbf{H}_u|\). The calculation of distance will be performed on this set, which can be seen as a base of a vector space.

Now we introduce the method for calculating distance.
To better calculate the distance between different sequences, Wasserstein distance \cite{panaretos2019statistical} has emerged as an effective method. The Wasserstein \(p\)-distance between two probability measures \(\mu\) and \(\nu\) is expressed as
\begin{equation}
\begin{split}
    W_p(\mu,\nu)=\inf_{\gamma\in\Gamma(\mu,\nu)} \left(\iint_{(x,y)}\gamma(x,y)[d(x,y)]^p\mathrm{d}x\mathrm{d}y\right)^{1/p} \\
    \text{s.t.} \quad \int \gamma(x,y)\mathrm{d}y=\mu(x), \int \gamma(x,y)\mathrm{d}x=\nu(y),
\end{split}
\end{equation}
where \(\gamma\) is a joint probability and \(d\) is the distance function of the metric space.
The Wasserstin-Fourier (WF) Distance \cite{cazelles2020wasserstein} between two discrete time series \(x\) and \(y\) of the same length can be represented as \(\text{WF}(x,y)=W_2(s_x,s_y)\), in which \(s(x)\) is the normalized power spectral density. The process of signal transformation can be expressed by the following flow:
\[x(t)\xrightarrow{\text{FT}}\hat{x}(f)\rightarrow S(f)\rightarrow s(f).\]
For discrete time series, the spectral density is calculated based on the frequency domain representation \(\hat{x}(f)\), where \(S(f_i) = |\hat{x}(f_i)|^2\), and the normalized spectral density is given by \(s(f_i) = \frac{S(f_i)}{\sum_i S(f_i)}\). Inspired by the Wasserstein-Fourier distance, we develop a simpler distance metric that uses the difference of time series in the frequency domain \(\hat{x}(f)\) to measure the distance between two time series.

We can calculate Wasserstein distance between \(\Lambda_{i,\mu}(\mathbf{H}_u)\) and \(\Lambda_{j,\mu}(\mathbf{H}_u)\) by:
\begin{equation}
    W_p(v_i,v_j)=\inf_{\pi:\mathbb{R}\rightarrow\mathbb{R}}\left(\frac{1}{T_u}\sum_{k\in \mathbf{H}_u}
\left|\Lambda_{i,\mu}(k)-\Lambda_{j,\mu}(\pi(k))\right|
^p\right)^{1/p}.
\end{equation}

In order to retain our intention of the reduction of complexity, we let \(\pi(k)=k\) and define \textbf{Fourier Sparse Distance} as: (take p as 2)
\begin{equation}
    \text{FSD}(v_i,v_j)=\sqrt{\frac{1}{T_u}\sum_{k\in \mathbf{H}_u}
\left|\Lambda_{i,\mu}(k)-\Lambda_{j,\mu}(k)\right|
^2}.
\end{equation}

We obtain a distance matrix \(\mathbf{A}^{N\times N}\) by calculating FSD between each pair of nodes. After obtaining the distance metric, for each node, we select the topk
k nodes with the smallest distances as its neighbor nodes. For these neighbor nodes, we set the values in the dynamic spatiotemporal graph to the normalized distance, indicating their relationships. For non-neighbor nodes, we set the values in the dynamic spatiotemporal graph to -inf. In this way, we obtain the dynamic mask matrix \(Mask_{dyn}\) .

\subsection{Modeling the Spatiotemporal Relations on Server}

 
First, we define two special attention masks. To model the static spatial relationships among the nodes, we define the distance mask as \(Mask_{dis}\) where the element in it is calculated based on distance relationships. The specific calculation method is detailed in the Experimental Setup section. In order to model the dynamic relations among nodes, we use the adaptive spatiotemporal aware graph construction module mentioned in the previous section to construct \(Mask_{dyn}\) from trend features \(\{trend_1,\cdots,trend_N\}\).

Then, we propose a spatiotemporal encoder layer based on specific masked spatial attention to merge the temporal features \(H = \{h_i,\cdots,h_N\} \) obtained from 
$N$ clients which contains the temporal dynamics of each client. 
Traditional attention forecasting modules apply self-attention operations in the temporal dimension, calculating relations between different time steps. In contrast, our method uses an attention module to model the relationships between nodes at different locations within the same time period.
Mask attention has been introduced to time series forecasting in PDFormer \cite{jiang2023pdformer}, and can be represented as follows:
\begin{equation}
   Attention_{i} = \mathrm{Softmax}(\frac{{Q_{i}} {K_{i}}}{\sqrt{d}} \odot Mask) \ {V_{i}}^S,
\end{equation}
where \(\odot\) is the Hadamard product. The elements of the mask matrix are either 0 or 1. This mask attention is a special case of the traditional mask fill which set mask element to $-\infty$. Since most elements of the mask matrix are set to 0, we have found in practice that directly using the mask does not significantly degrade performance. So we optimize the critical \(Q \cdot K\) computation step of the attention module to further reduce computational costs to reduce computational overhead and conform to the computational resource constraints of edge devices. We hypothesize that the attention map obtained from \( Q \cdot K \cdot Mask \) can be approximated by the spatiotemporal relationship \(Mask\).
This approach forces the attention mechanism to focus more on the pre-modeled static spatial relationships and dynamic spatio-temporal relationships. Finally, we can define the output from the spatial attention modules as:

\begin{equation}
   Attention_{i} = \mathrm{Softmax}(Mask) \ {V_{i}}^S,
\end{equation}
where \(Mask \in \{Mask_{dis},Mask_{dyn}\}, \ {V_{i}}^S = H W_{V_{i}} \). \(W_{V_{i}}\) are learnable parameters.

After merging the spatial-temporal information using mask spatial attention blocks, we obtained \(h_{static}\) and \(h_{dynamic}\) by aggregating client temporal features based on the static mask and dynamic mask. We concate two hidden values and transfer the aggregated message \(h_{agg}\) to clients for the prediction process.


\subsection{Privacy Analysis}
\subsubsection{Attack Model} 



 We assume a curious but benign server managed by an \textit{authorized organization}. Most participants do not tamper with data or upload fake messages that could disrupt the system. However, \textit{attackers} may be among users or within the server. An attacker may also obtain information by eavesdropping on the communication channel.
\textit{Client side}:  
Each client only shares local temporal and trend features with the server, then downloads aggregated features for prediction. This ensures the security of the client.
\textit{Server side}: 
As discussed, the server and users exchange thresholded Fourier components and the final states of the encoder.  The server may perform reverse engineering such as inverse Fourier transform, attempting to recover time series data utilizing features uploaded by the client. Subsequently, the server-side may utilize the reconstructed data to train a predictive model aimed at forecasting the user's future privacy conditions. For instance, the model could predict whether the user's electricity consumption is zero to infer whether the user is at home.

\subsubsection{Protect Method}
Above widely used Gaussian mechanism and Laplace mechanism \cite{wei2020federated}, we further propose a more generalized mechanism which allows more personalised privacy requirement.
Consider stable noise \(\mathbf{n}\in\mathbb{R}^{k}\) which has characteristic function \(\phi(t;\alpha,\beta,c,\mu)\), and the probability density function is
\[f(x)=\frac{1}{2\pi}\int\phi(t)e^{-ixt}\,dt.\]
\(\alpha=2\) leads to normal distribution which has be discussed. The distributions with \(0<\alpha<2\) has heavier tails and infinite variance, sampled from which results in more outliers. Considering the attacker does not have the knowledge of the noise generator, a large amount of outliers can produce better confusion, thereby reducing the success rate of the attack. Especially, \(\alpha=1\) leads to Cauchy distribution.
We adopt differential privacy (DP) to quantify the effectiveness of proposed mechanism. Detailed introduction of DP is provided in Appendix A.

\begin{theorem}[Constraints of Cauchy mechanism]
A Cauchy mechanism with scale factor \(c>\frac{\Delta f(1+e^{\epsilon/2})}{(e^\epsilon-1)(\frac{1}{2}+\tan(\pi\delta))}\) provides \((\epsilon,\delta)\)-DP.
\end{theorem}
\begin{proof}
\vspace{-4pt}
    See Appendix C.
\vspace{-8pt}
\end{proof}

In practical, each client encodes the seasonal term \(\mathbf{x}_{i,0}^{(r)}\) using \(f_1\) during their local training. At the end of each global communication round \(r\), each client generate an stable noise vector and add it to the resulting states, which can be formulated as \(\Tilde{\mathbf{h}}_i^{(r)}=\mathbf{h}_i^{(r)}+\mathbf{n}_{i,0}^{(r)}=f_1(\mathbf{x}_{i,0}^{(r)})+\mathbf{n}_{i,0}^{(r)}\).
For simplicity, each element of \(\mathbf{n}_{i,0}^{(r)}\) follows the same distribution.
On the other hand, at the beginning of each period of dynamic graph construction, users also add a stable noise to the results of filter Fourier transform \(f_2\) performing on the trend term \(\mathbf{x}_{i,1}^{(r)}\), which can be formulated as \(\Tilde{\Lambda}_{i,\mu}^{(r)}= \Lambda_{i,\mu}^{(r)}+\mathbf{n}_{i,1}^{(r)}= f_2(\mathbf{x}_{i,1}^{(r)})+\mathbf{n}_{i,1}^{(r)}\). For each element of \(\mathbf{n}_{i,1}^{(r)}\), both \(\beta\) and shift factor \(\mu\) are set to zero, and the scale factor is set to \(c=E\sigma(\Lambda_{i,\mu}(k))\), in which \(E\ge 0\) is the noise intensity and \(\sigma:\mathbb{C}\rightarrow\mathbb{R}\) is the scaling function. Notice that the mechanism only changes amplitude over fixed frequency without changing the phase and will not introduce any new frequency components.


\section{Experiment}

\subsection{Experiment Settings}
\begin{table}[]
\caption{Statistics of datasets}
\centering
\small
\begin{tabular}{cccc}
\toprule
Dataset & \# Nodes & \# Edges & \# Time Steps \\ \midrule
PEMS03  & 358 & 304 & 26,208      \\
PEMS04  & 307 & 340 & 16,992        \\
PEMS08  & 170 & 295 & 17,856         \\
METR-LA & 270 & 1515 & 34,272         \\
Solar & 137 & null & 52,560 \\
ECL & 321 & null & 26,304 \\ \bottomrule
\end{tabular}
\label{tab:data}
\vspace{-10pt}
\end{table}
\textbf{Datasets}: We evaluate our proposed FedASTA method on typical spatiotemporal forecasting tasks. We use dataset from three different domains: traffic flow, solar system, and electricity, all of which have privacy concerns mentioned in previous sections.

In these datasets, the traffic flow dataset provides access to distance information between sensors, which we use to create a static distance graph between the sensors. However, for the solar and electricity datasets, we cannot obtain the location information of the sensors. Therefore, we set the static distance graph for these two datasets as an identity matrix.
Their statistics are shown in Table \ref{tab:data}.

\noindent \textbf{Baselines}: To comprehensively test the performance of our method, we select several popular federated learning methods in modeling spatiotemporal data as baselines.

The baselines can be devided into three classes. (1) \textit{Traditional Federated learning methods}: We choose FedAvg \cite{mcmahan2017communication}, FedProx \cite{li2020federated} and FedRep \cite{pmlr-v139-collins21a}, which are the most general federated learning algorithms. (2) \textit{Attention based methods}: We choose FedAtt \cite{8852464} and FedDA \cite{9488883}. These methods  use attentive aggregation module that considers the contributions of client models to the global model. (3) \textit{GNN based method}: We choose CNFGNN \cite{meng2021cross} which employs an encoder-decoder structure to capture temporal dynamics on each client node and a graph neural network (GNN) on the server node to model spatial dynamics.

\noindent \textbf{Other settings:}
To ensure fair comparison, we split the data into training, validation, and test sets as done in CNFGNN. We use one hour of historical data to predict the next hour's traffic flow. For static graph-based methods, we construct a static distance graph for each dataset as described in \cite{meng2021cross}. Datasets without sensor location information use an identity matrix for their static graph. To avoid losing continuous temporal features when shuffling the training data, we adopt the graph construction method from DSTAGNN \cite{lan2022dstagnn}, pre-building dynamic spatiotemporal relationship graphs for different time periods. Our graph construction module only generates mask matrices, eliminating data leakage risk.

For client models in the federated setting, we use a two-layer GRU with a hidden dimension of 100 and an Avgpool window size of 5. The hidden dimensions of server models for both CNFGNN and our model are set to 200, with 2 encoder layers on our server model. All models are trained for 100 rounds. Model performance is evaluated using mean absolute error (MAE), mean absolute percentage error (MAPE), and root mean squared error (RMSE).  
\subsection{Results and Analyses}

\begin{table*}
    \centering
    \caption{Experimental Results. All methods except our proposed method are reimplemented. We use Mean Absolute Error (MAE), Mean Absolute Percentage Error (MAPE) and Root Mean Square Error (RMSE) as metrics.}
    \resizebox{\linewidth}{!}{
    \Huge 
    \renewcommand{\arraystretch}{1.5} 
    \begin{tabular}{ccccccccccccccccccc}
    \toprule
    \multirow{2}{*}{Methods} & \multicolumn{3}{c}{METR-LA} & \multicolumn{3}{c}{PEMS03} & \multicolumn{3}{c}{PEMS04} & \multicolumn{3}{c}{PEMS08} & \multicolumn{3}{c}{Solar} & \multicolumn{3}{c}{ECL} \\ \cmidrule(lr){2-4} \cmidrule(lr){5-7} \cmidrule(lr){8-10} \cmidrule(lr){11-13} \cmidrule(lr){14-16} \cmidrule(lr){17-19}
                             & MAE     & MAPE    & RMSE    & MAE     & MAPE    & RMSE   & MAE     & MAPE    & RMSE   & MAE     & MAPE    & RMSE   & MAE    & MAPE    & RMSE   & MAE    & MAPE   & RMSE  \\ \midrule
    FedAvg                   & 9.15    & 13.67   & 12.81   & 24.75   & 23.17   & 40.17  & 29.68   & 20.06   & 46.21  & 24.66   & 22.63   & 36.87  & 2.28   & 46.29   & 3.77   & 0.54   & 58.75  & 0.75  \\
    FedProx                  & 6.21    & 13.21   & 12.43   & 24.78   & 23.25   & 39.98  & 29.23   & 19.86   & 44.59  & 23.46   & 18.96   & 35.35  & 2.15   & 43.78   & 3.72   & 0.53   & 57.15  & 0.73  \\
    FedAtt                   & 6.82    & 14.34   & 12.29   & 23.66   & 20.83   & 38.76  & 29.59   & 19.30   & 44.60  & 24.19   & 21.92   & 36.36  & 2.13   & 45.83   & 3.71   & 0.51   & 55.17  & 0.73  \\
    FedDA                  &  6.61   &  13.81  &  12.17  &  21.95  &  27.81  &  34.57 &  29.29  & 20.18   & 44.18  & 23.37   & 19.39   &  34.85 & 2.26  &  43.44  &  3.67  & 0.51   & 55.65  & 0.75  \\
    FedRep                   & 7.17    & 13.55   & 12.67   & 24.32   & 23.13   & 39.70  & 29.79   & 29.17   & 43.25  & 25.61   & 24.32   & 36.93  & 1.94   & 45.41   & 3.80   & 0.54   & 51.41  & 0.75  \\
    CNFGNN                   & 5.73    & 11.40   & 11.79   & 23.21   & 24.48   & 31.94  & 29.78   & 18.64   & 41.46  & 24.12   & 28.10   & 33.83  & 1.97   & 44.37   & 3.57   & 0.55   & 56.85  & 0.73  \\
    FedASTA                  & \textbf{5.70} & \textbf{11.31} & \textbf{11.60} & \textbf{17.62} & \textbf{19.81} & \textbf{29.51} & \textbf{20.34} & \textbf{14.22} & \textbf{32.55} & \textbf{16.76} & \textbf{12.25} & \textbf{26.66} & \textbf{1.72} & \textbf{42.91} & \textbf{3.26} & \textbf{0.42} & 57.01 & \textbf{0.65}
    \\ \bottomrule
    \end{tabular}
    }
    \label{tab:main_result}
\vspace{-10pt}
\end{table*}
The comparison results with baselines on all datasets are shown in Table \ref{tab:main_result}
. We can clearly see that our method has achieved the best performance on all datasets under the federated learning settings. We can draw the following observations: (1) General federated learning methods such as FedAvg and FedProx exhibit significant accuracy degradation as they fail to capture the spatial-temporal dynamics in the data. (2) While CNFGNN considers static distance relations between different nodes, it falls short in capturing the dynamic spatial relations between different clients. (3) Our method effectively captures the temporal-spatial dynamics, resulting in superior performance compared to these methods. We also present the plots of the prediction results and ground truth curves for different methods over a single day in Appendix G.

\subsection{Model Scalability Study}

\begin{figure}
    \centering
    \includegraphics[width=\linewidth]{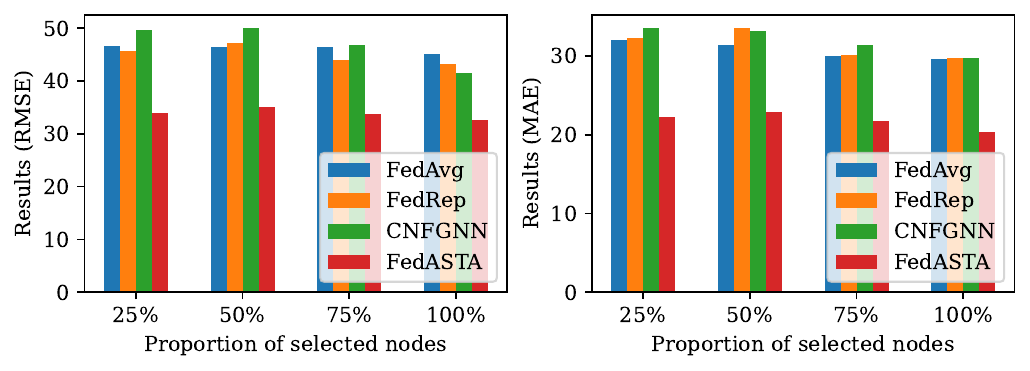}
    \caption{Performance of the model under different numbers of nodes and edges (left, RMSE; right, MAE)}
    \label{fig:Model Scalarbility}
\vspace{-10pt}
\end{figure}

In this section, we analyze the scalability of our model by evaluating its predictive performance under varying numbers of nodes and predefined static edges. Using the PEMS04 dataset, we created four sub-datasets with 76, 153, 230, and 307 nodes and corresponding edge counts of 97, 218, 347, and 516, respectively. We compared our method against three baselines: FedAvg, FedRep, and CNFGNN, using RMSE as the evaluation metric. The results are shown in Fig. \ref{fig:Model Scalarbility}.

Our observations indicate that as the number of nodes and edges increases, prediction error for all models decreases, likely due to more accurate temporal feature estimation. FedAvg and FedRep, which don't rely on a static graph, are less sensitive to changes in node and edge counts. CNFGNN, which uses a static graph, performs worse with fewer edges due to its inability to capture dynamic dependencies. Our method, considering both static and dynamic spatiotemporal dependencies, demonstrates excellent scalability and remains robust to changes in nodes and edges.

\subsection{Communication Efficiency Study}
In this section, we analyze the communication cost during the model training processes in each communication round. 

During training, traditional methods like FedAvg, FedRep, and FedDA require uploading local model parameters to the server after each round. In contrast, our method only merges hidden variables, with nodes uploading data to the server and receiving data back. Additionally, gradients are exchanged twice per batch, leading to significant communication overhead when multiplied by the length of the training data.

To address this, we adopted the two-stage training strategy from \cite{meng2021cross}, which trains edge node models and the server separately, greatly reducing overhead. The results, shown in Table \ref{tab:communication}, demonstrate that this approach significantly lowers training costs, making it comparable to traditional methods.

\begin{table}[htbp]
    \centering
    \caption{The communication overhead of different methods in one round of training.}
    \setlength{\tabcolsep}{4pt} 
    \renewcommand{\arraystretch}{1.2} 
    \resizebox{\linewidth}{!}{ 
    \begin{tabular}{lccc}
    \toprule
    Method Name & Hyperparameter & Metric \\
    \midrule
    \multirow{4}{*}{Merging parameters} & Node Number & \multicolumn{2}{c}{307} \\
                              & Node Model Weights Size (MB) & \multicolumn{2}{c}{1.15} \\
                              & Parameter Number & \multicolumn{2}{c}{300300} \\
                              & Train Comm Cost (MB) & \multicolumn{2}{c}{353.05} \\
    \midrule
    \multirow{4}{*}{Merging variables}  & Node Number & \multicolumn{2}{c}{307} \\
                              & Hidden State Size (MB) & \multicolumn{2}{c}{0.024} \\
                              & Train Data Length & \multicolumn{2}{c}{11872} \\
                              & Train Comm Cost (MB) & \multicolumn{2}{c}{11250} \\
    \midrule
    \multirow{5}{*}{Merging with two stages} & Node Number & \multicolumn{2}{c}{307} \\
                              & Node Model Weights Size (MB) & \multicolumn{2}{c}{1.15} \\
                              & Hidden State Size (MB) & \multicolumn{2}{c}{0.024} \\
                              & Total Rounds & \multicolumn{2}{c}{2} \\
                              & Train Comm Cost (MB) & \multicolumn{2}{c}{801.525} \\
    \bottomrule
    \end{tabular}
    }
    \label{tab:communication}
\vspace{-10pt}
\end{table}

\subsection{Ablation Study}
To verify the rational business of FedASTA components, we provide detailed ablations with removing components (w/o) experiments:
(1) \textbf{w/o time series decomposition:} this variant removes the decomposition part on the client.
(2) \textbf{w/o adaptive graph:} The model on the client is unchanged. Only dynamic graph attention within server's model are removed.
(3) \textbf{w/o static graph:} The model on the client is unchanged. Only static graph attention within server's model are removed.
(4) \textbf{w/o all graph:} The model on the client remains unchanged. Both dynamic graph attention and static graph attention are removed, leaving only the serial connection of the encoder, which contains a normal spatial attention module.


The results are presented in Table \ref{tab:ablation}. From the experimental results, we can draw the following conclusions: (1) In the absence of the decomposition module on the client nodes, extracting the trend from the complex original time series leads to a performance decline, highlighting the necessity of the time series decomposition module.
(2) The performance degrades significantly when both graphs are absent. This demonstrates the effectiveness of our proposed masked attention and graph construction modules. (3) On the dataset such as ECL dataset, which lacks predefined static graph relationships, dynamically modeling spatiotemporal relationships has a more significant impact on prediction performance. This further underscores the necessity of dynamically modeling spatiotemporal dependencies.
\begin{table}[]
\caption{Ablation study on the time series decomposition module and two spatial attention modules.}
\Huge 
\resizebox{\linewidth}{!}{
\begin{tabular}{ccccccc}
\toprule
\multirow{2}{*}{Methods}
                           & \multicolumn{2}{c}{PEMS03} & \multicolumn{2}{c}{PEMS04} & \multicolumn{2}{c}{Solar} \\ \cmidrule(lr){2-3} \cmidrule(lr){4-5} \cmidrule(lr){6-7}
                           & RMSE         & MAE         & RMSE         & MAE         & RMSE        & MAE         \\
                           \midrule
w/o Decomp.                & 31.24        & 18.27       & 34.42        & 23.16       & 3.56        & \textbf{1.45}        \\
w/o static graph           & 30.36        & 18.84       & 33.45        & 22.53       & 3.73        & 1.61        \\
w/o adapt. graph           & 30.52        & 17.83       & 33.90        & 22.86       & 3.84        & 1.68        \\
w/o all graph              & 33.58        & 19.92       & 37.17        & 24.11       & 4.26        & 1.89        \\
FedASTA                    & \textbf{29.51}        & \textbf{17.62}       & \textbf{32.55}        & \textbf{20.34}       & \textbf{3.26}        & 1.72       \\
\bottomrule
\end{tabular}
}
\label{tab:ablation}
\vspace{-15pt}
\end{table}

\subsection{Evaluation of Privacy Mechanism}
\begin{figure}[htbp]
    \centering
    \begin{subfigure}{0.52\linewidth}
        \centering
        \includegraphics[width=\textwidth]{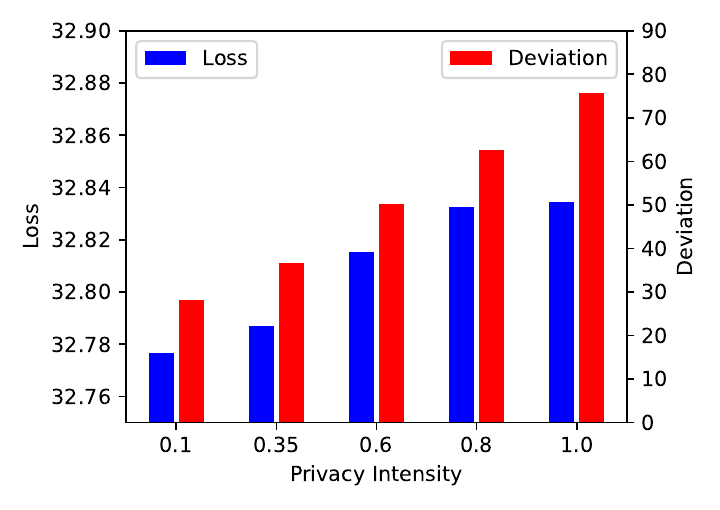}
        \caption{Effects of Noise Intensity on Model Performance and Reconstruction Error.}
        \label{fig:loss}
    \end{subfigure}\hfill
    \begin{subfigure}{0.46\linewidth}
        \centering
        \includegraphics[width=\textwidth]{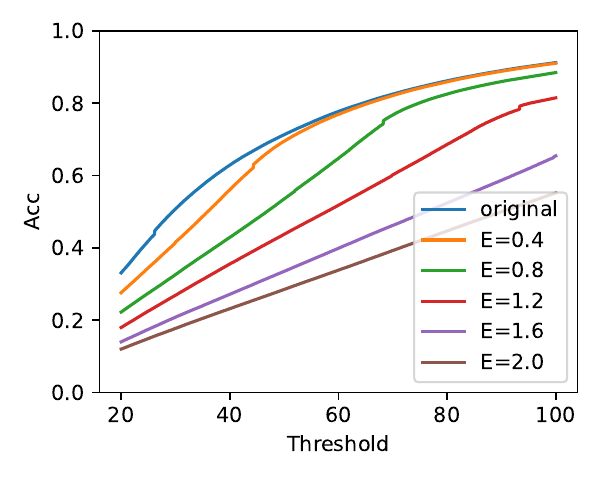}
        \caption{Relationship Between Reconstruction Attack Error and Noise Intensity.}
        \label{fig:bias}
    \end{subfigure}
    \vspace{-6pt}
    \caption{Effect of privacy intensities.}
    \label{fig:combined}
    \vspace{-16pt}
\end{figure}

As discussed in the section on the attack model, we conducted two privacy analysis experiments. The first experiment examined the relationship between the magnitude of added noise and the mean squared error between the recovered and original sequences. In the second experiment, we trained a GRU model using the reconstructed data to predict the user's future time series data, such as electricity consumption. The reconstructed data were obtained by restoring trend features via inverse Fourier transform. A threshold was set, defining a successful attack as one where the predicted value falls within this threshold relative to the true value. As shown in Figure \ref{fig:loss}, both the model's predictive performance and the error in the directly recovery increase with the level of noise. Despite being affected by noise, the model remains relatively robust. The attack accuracy varies with the threshold, as shown in Figure \ref{fig:bias}. When privacy intensity is sufficiently high, even with loose constraints on attack success, achieving a 50\% success rate becomes challenging for the attacker.


\subsection{Case Study}
\begin{figure}
    \centering
    \includegraphics[width=\linewidth]{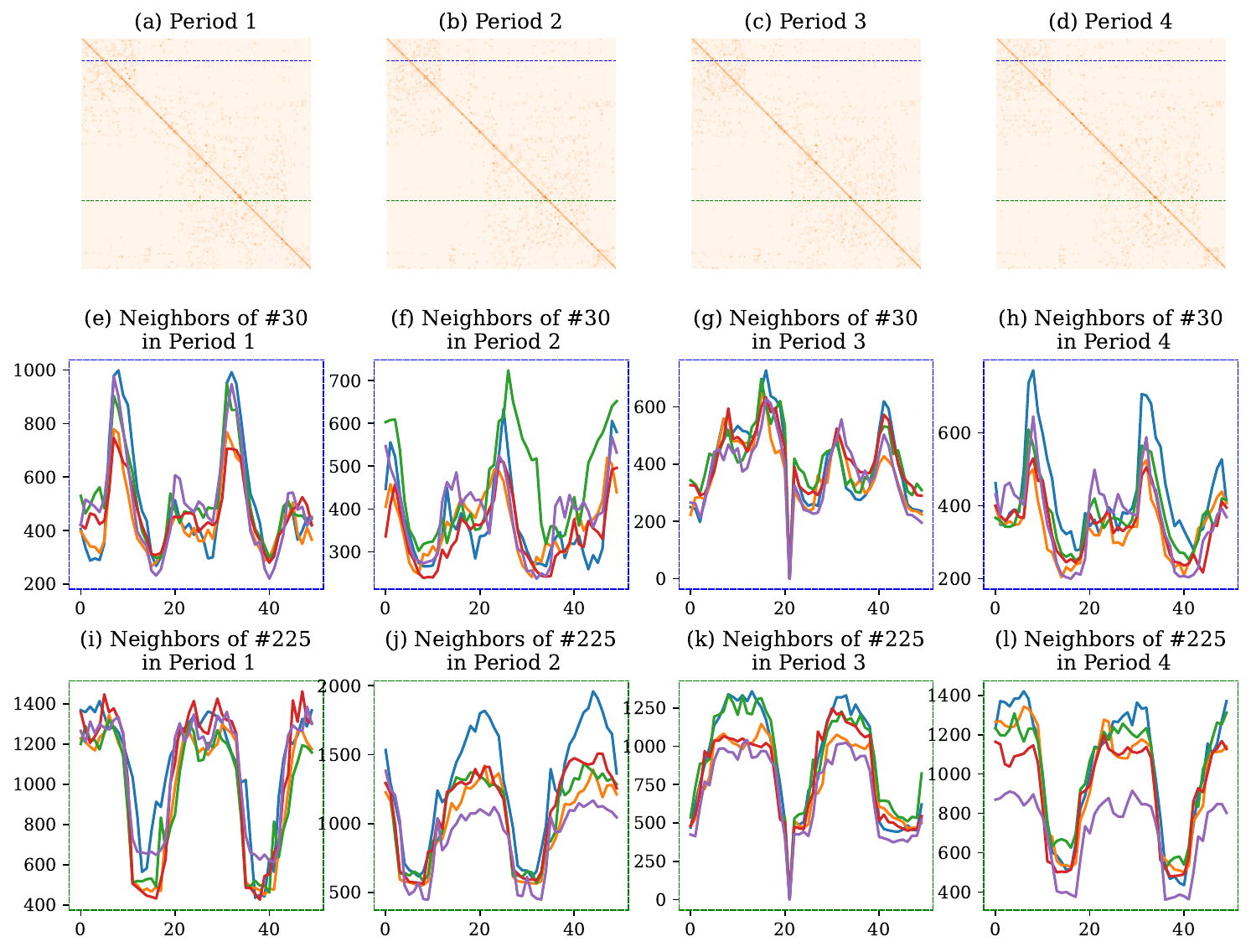}
    \caption{Effectiveness of dynamic graph construction mechanism. (a)-(d) shows the attention maps of 4 disjoint periods. The darker color represents stronger correlations. The blue line and green line represent the similarity vector of nodes \#30 and \#225 respectively. (e)-(h) and (i)-(l) shows the different periods from 4 neighbors of nodes \#30 and \#225 and themselves respectively.}
    \label{fig:34}
\vspace{-16pt}
\end{figure}

\begin{table}[]
    \caption{Influence of the number of edges}
    \centering
    \scalebox{0.85}{
    \begin{tabular}{ccccccc}
    \toprule
\multirow{2}{*}{Metric} & \multicolumn{6}{c}{\#edges} \\ \cmidrule(lr){2-7} 
                        & 1     & 8     & 16    & 32    & 64    & 128   \\
                        \midrule
RMSE                    & 33.13 & 32.67 & 32.65 & \underline{32.61} & 33.17 & 33.69 \\
MAE                     & 21.4  & 21    & 21.02 & \underline{20.95} & 21.55 & 21.01 \\
MAPE (\%)                & 15.55 & 14.87 & 15.38 & \underline{14.79} & 16.56 & 14.88 \\
    \bottomrule
    \end{tabular}
    }
    \label{tab:edg}
\vspace{-12pt}
\end{table}
To enhance the interpretability of our model, we visualized the dynamic attention maps constructed at different time stages in the electricity dataset as shown in Figure \ref{fig:34}. We use node 30 (indicated by the blue line in the figure) and node 225 (indicated by the green line in the figure) as examples. It can be observed that our dynamic spatiotemporal relationship graph successfully captures the spatiotemporal relationships between nodes, with similar temporal features among neighboring nodes. Additionally, the temporal features of the same node differ across different time periods, further demonstrating the effectiveness and importance of dynamic graph construction.

In addition, we conducted experiments on the impact of varying the number of edges in the dynamic graph on performance.
The results are shown in Table \ref{tab:edg}. We can  that too few edges fail to accurately model correlations between nodes, while too many edges introduce nodes with dissimilar temporal features, reducing predictive performance. 
Thus, selecting an appropriate number of edges is crucial for our module.
 
\section{Conclusion}

We propose FedASTA, a novel federated learning (FL) framework for spatio-temporal data modeling. FedASTA includes an adaptive spatial-temporal graph construction module to capture complex inter-node relationships, and a masked spatial attention module that accounts for both static and dynamic spatial relations while reducing computational overhead.
Extensive experiments on six real-world datasets show that FedASTA outperforms other FL baselines. Additional tests demonstrate the effectiveness of each module in our model. We hope this approach offers new insights for spatio-temporal data modeling in FL scenarios.

\bibliographystyle{plain}
\bibliography{aaai25}

\newpage
\section*{Reproducibility Checklist}
This paper:

\begin{itemize}
    \item Includes a conceptual outline and/or pseudocode description of AI methods introduced (yes/partial/no/NA) \textcolor{red}{yes}
    \item Clearly delineates statements that are opinions, hypothesis, and speculation from objective facts and results (yes/no) \textcolor{red}{yes}
    \item Provides well marked pedagogical references for less-familiare readers to gain background necessary to replicate the paper (yes/no) \textcolor{red}{yes}
\end{itemize}

Does this paper make theoretical contributions? (yes/no) \textcolor{red}{yes}

If yes, please complete the list below.
\begin{itemize}
    \item All assumptions and restrictions are stated clearly and formally. (yes/partial/no) \textcolor{red}{yes}
    \item All novel claims are stated formally (e.g., in theorem statements). (yes/partial/no) \textcolor{red}{yes}
    \item Proofs of all novel claims are included. (yes/partial/no) \textcolor{red}{yes}
    \item Proof sketches or intuitions are given for complex and/or novel results. (yes/partial/no) \textcolor{red}{partial, the proof of the proposed mechanism is similar to existing methods}
    \item Appropriate citations to theoretical tools used are given. (yes/partial/no) \textcolor{red}{yes}
    \item All theoretical claims are demonstrated empirically to hold. (yes/partial/no/NA) \textcolor{red}{yes}
    \item All experimental code used to eliminate or disprove claims is included. (yes/no/NA) \textcolor{red}{yes}
\end{itemize}

Does this paper rely on one or more datasets? (yes/no) \textcolor{red}{yes}

If yes, please complete the list below.
\begin{itemize}
    \item A motivation is given for why the experiments are conducted on the selected datasets (yes/partial/no/NA) \textcolor{red}{yes}
    \item All novel datasets introduced in this paper are included in a data appendix. (yes/partial/no/NA) \textcolor{red}{NA, no novel datasets are introduced}
    \item All novel datasets introduced in this paper will be made publicly available upon publication of the paper with a license that allows free usage for research purposes. (yes/partial/no/NA) \textcolor{red}{NA, no novel datasets are introduced}
    \item All datasets drawn from the existing literature (potentially including authors’ own previously published work) are accompanied by appropriate citations. (yes/no/NA) \textcolor{red}{yes}
    \item All datasets drawn from the existing literature (potentially including authors’ own previously published work) are publicly available. (yes/partial/no/NA) \textcolor{red}{yes}
    \item All datasets that are not publicly available are described in detail, with explanation why publicly available alternatives are not scientifically satisficing. (yes/partial/no/NA) \textcolor{red}{NA, all datasets that are publicly available}
\end{itemize}

Does this paper include computational experiments? (yes/no) \textcolor{red}{yes}

If yes, please complete the list below.
\begin{itemize}
    \item Any code required for pre-processing data is included in the appendix. (yes/partial/no). \textcolor{red}{yes}
    \item All source code required for conducting and analyzing the experiments is included in a code appendix. (yes/partial/no) \textcolor{red}{yes}
    \item All source code required for conducting and analyzing the experiments will be made publicly available upon publication of the paper with a license that allows free usage for research purposes. (yes/partial/no) \textcolor{red}{yes}
    \item All source code implementing new methods have comments detailing the implementation, with references to the paper where each step comes from (yes/partial/no) \textcolor{red}{yes}
    \item If an algorithm depends on randomness, then the method used for setting seeds is described in a way sufficient to allow replication of results. (yes/partial/no/NA) \textcolor{red}{yes}
    \item This paper specifies the computing infrastructure used for running experiments (hardware and software), including GPU/CPU models; amount of memory; operating system; names and versions of relevant software libraries and frameworks. (yes/partial/no) \textcolor{red}{yes}
    \item This paper formally describes evaluation metrics used and explains the motivation for choosing these metrics. (yes/partial/no) \textcolor{red}{yes}
    \item This paper states the number of algorithm runs used to compute each reported result. (yes/no) \textcolor{red}{yes}
    \item Analysis of experiments goes beyond single-dimensional summaries of performance (e.g., average; median) to include measures of variation, confidence, or other distributional information. (yes/no) \textcolor{red}{yes}
    \item The significance of any improvement or decrease in performance is judged using appropriate statistical tests (e.g., Wilcoxon signed-rank). (yes/partial/no) \textcolor{red}{yes}
    \item This paper lists all final (hyper-)parameters used for each model/algorithm in the paper’s experiments. (yes/partial/no/NA) \textcolor{red}{partial, some commonly used (hyper-) parameters (e.g., learning rate) are not listed in this paper}
    \item This paper states the number and range of values tried per (hyper-) parameter during development of the paper, along with the criterion used for selecting the final parameter setting. (yes/partial/no/NA) \textcolor{red}{partial, some commonly used (hyper-) parameters (e.g., learning rate) are not explained in this paper}
\end{itemize}

\end{document}